\documentclass[runningheads]{llncs}

\usepackage{eccv}

\usepackage{eccvabbrv}

\usepackage{graphicx}
\usepackage{booktabs}

\usepackage[accsupp]{axessibility}  

\usepackage{graphicx}
\usepackage{amsmath}
\usepackage{amssymb}
\usepackage{multirow}
\usepackage{multicol}
\usepackage{color}
\usepackage{wrapfig}
\usepackage{tcolorbox}

\PassOptionsToPackage{table,dvipsnames}{xcolor}

\usepackage[table]{xcolor}
\usepackage[dvipsnames]{xcolor}
\usepackage{colortbl}
\usepackage{xspace}
\usepackage{mdframed}
\usepackage{comment}

\definecolor{mygray}{RGB}{230,230,230}
\definecolor{eccvblue}{rgb}{0.30,0.49,0.85}
\definecolor{linkpink}{RGB}{252,90,122}
\definecolor{linkgray}{RGB}{50,50,50}
\usepackage[colorlinks,bookmarksopen,citecolor=eccvblue, urlcolor=magenta]{hyperref}

\usepackage{hyperref}

\usepackage{orcidlink}

\newcommand{\tablestyle}[2]{\setlength{\tabcolsep}{#1}\renewcommand{\arraystretch}{#2}\centering\small}

\begin{document}

\title{Can OOD Object Detectors Learn from Foundation Models?}


\author{\small Jiahui Liu \and Xin Wen \and Shizhen Zhao \and Yingxian Chen \and Xiaojuan Qi\thanks{corresponding author}}

\authorrunning{Liu et al.}

\institute{\small The University of Hong Kong \\
\email{\{liujh,wenxin,zhaosz,chenyx,xjqi\}@eee.hku.hk} \\ \url{https://github.com/CVMI-Lab/SyncOOD}
}

\maketitle

\begin{abstract}
Out-of-distribution (OOD) object detection is a challenging task due to the absence of open-set OOD data. Inspired by recent advancements in text-to-image generative models, such as Stable Diffusion, we study the potential of generative models trained on large-scale open-set data to synthesize OOD samples, thereby enhancing OOD object detection. We introduce SyncOOD, a simple data curation method that capitalizes on the capabilities of large foundation models to automatically extract meaningful OOD data from text-to-image generative models. This offers the model access to open-world knowledge encapsulated within off-the-shelf foundation models. The synthetic OOD samples are then employed to augment the training of a lightweight, plug-and-play OOD detector, thus effectively optimizing the in-distribution (ID)/OOD decision boundaries. Extensive experiments across multiple benchmarks demonstrate that SyncOOD significantly outperforms existing methods, establishing new state-of-the-art performance with minimal synthetic data usage. 

\end{abstract}

\section{Introduction}
\label{sec:intro}

\begin{figure}[t]
  \centering
  \includegraphics[width=\textwidth]{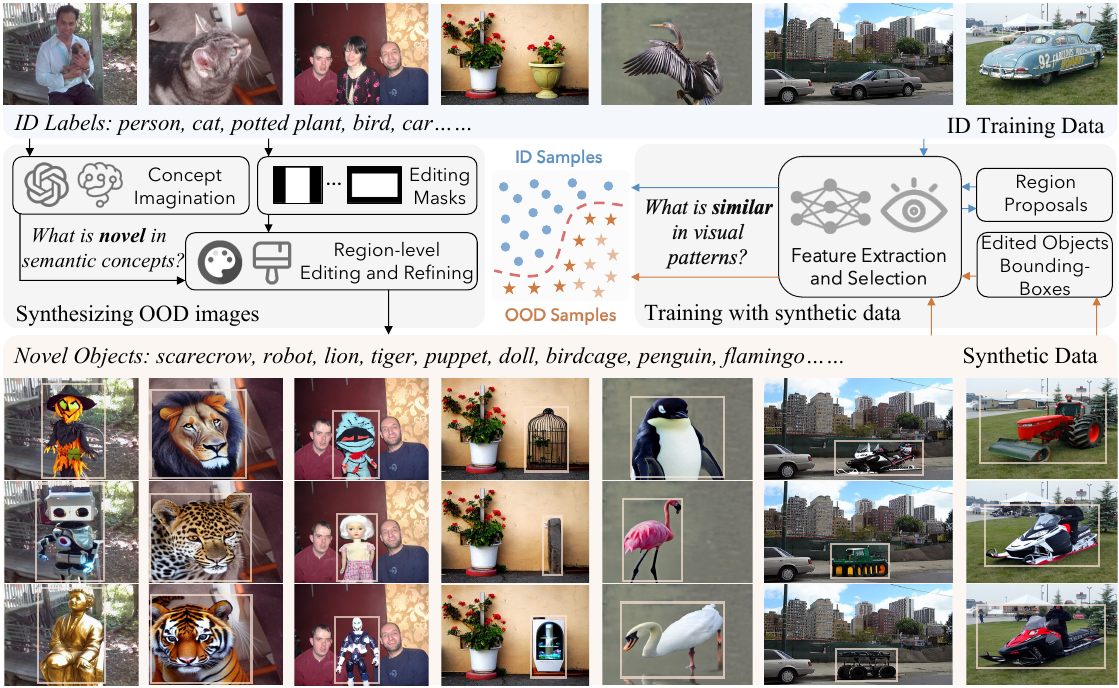}
  \vspace{-0.5cm}
  \caption{Our pipeline replaces ID objects with semantic-novel yet visual-similar objects for scene-level OOD object synthesis. Middle left: The concepts are imagined by an LLM to ensure semantic separability and rationality, and reformed as text prompts for controllable in-painting using Stable Diffusion. Middle right: During training, only visually similar OOD objects are adopted based on instance-level feature similarity to the original object. A lightweight binary classifier is optimized for OOD detection, and other parts of the detector are kept unchanged.}
  \vspace{-0.5cm}
  \label{fig:main}
\end{figure}

Modern object detectors, trained on closed-set data, have achieved remarkable success. However, they often incorrectly and confidently classify out-of-distribution (OOD) object categories as in-distribution (ID) categories in open-world applications~\cite{dhamija2020overlooked}, raising concerns about their reliability for deployment. To enhance the trustworthiness of object detection, researchers have studied the OOD object detection task, which aims to identify and flag unknown or novel objects as distinct from ID ones~\cite{bishop1994novelty,nguyen2015deep}.

The vulnerability of these models to OOD samples stems from their lack of awareness of the unknown open data distribution during training. Consequently, synthesizing OOD samples for model learning has emerged as a major research direction for this task~\cite{du2022vos,wu2023deep,wu2023discriminating,wilson2023safe}. Most existing studies~\cite{du2022vos,wu2023deep,wu2023discriminating} concentrate on generating OOD objects in the latent space of an object detection model trained on ID data. These synthesized samples are then used to optimize the decision boundary between ID and OOD data. Alternative approaches include directly synthesizing images by injecting adversarial noise~\cite{wilson2023safe} or identifying OOD instances from video data~\cite{du2022unknown}. While these methods have yielded promising results, they remain limited to a closed-set setting, where the latent space for synthesizing outliers or the data is derived from a closed-set data distribution. Consequently, they may be biased towards the ID dataset, leading to suboptimal performance. Besides, effectively handling the unknown may appear unattainable when the unknown is never fully exploitable. Beyond that, \textit{is it possible to learn them from massive open-world data knowledge condensed in foundation models?}

To this end, we investigate text-to-image generation models trained on a large amount of open-set data, which have demonstrated a superior ability to capture the distribution of data across a wide range of visual concepts, in order to synthesize novel data samples for enhancing OOD detection. Nonetheless, automatically extracting meaningful data from generative models for OOD object detection remains challenging due to the extensive vocabulary space to be explored, the complex scene-level synthesis problem, the need for object-level annotations, and potential distractions from contextual cues.

We introduce SyncOOD, an automatic data curation process that leverages foundation models as tools to harvest meaningful data from text-to-image generation models for OOD object detection (see Figure \ref{fig:main}).
The process is based on two key observations: 1) Hard OOD samples that are close to the ID data contribute more to learning a better OOD detector, and 2) Context may become a distracting cue for OOD object detection tasks, leading to bias towards contexts. With these observations in mind, the outlier synthesis process is formulated as box-conditioned image in-painting, and driven by Stable Diffusion~\cite{rombach2022high} for high-quality controllable editing. The concepts to replace with are imagined by a Large Language Model (LLM)~\cite{achiam2023gpt} with the aim of semantic novelty,\footnote{Concepts overlapping with the test data are removed to avoid information leakage.} and the associated bounding box is further refined with SAM~\cite{kirillov2023segment}. Automated by foundation models, this data collection pipeline requires minimal human labor, while producing high-quality OOD data.

In comparison to existing methods for OOD object detection,  our core insight is to broaden the model's exposure to a more extensive range of open-set data and circumvent dataset biases by tapping into the open-world knowledge found in off-the-shelf foundation models. Utilizing generative models also provides us with control over the context of synthesized images and the data distribution. Our comprehensive experiments demonstrate superior performance, emphasizing the untapped potential of text-image-generation models in the context of OOD object detection. Our key contributions are summarized as follows:

\begin{itemize}

\item We investigate and unlock the potential of text-to-image generative models trained on large-scale open-set data for synthesizing OOD objects in object detection tasks. 

\item We introduce an automated data curation process for obtaining controllable, annotated scene-level synthetic OOD images for OOD object detection, which utilizes LLMs for novel concept discovery and visual foundation models for data annotation and filtering.

\item We discover that maintaining ID/OOD image context consistency and obtaining more accurate OOD annotation bounding boxes are crucial for synthesized data to be effective in OOD object detection.

\item Comprehensive experiments on multiple benchmarks demonstrate the effectiveness of our method, as we significantly outperform existing state-of-the-art approaches while using minimal synthetic data.

\end{itemize}

\section{Related Work}
\label{sec:relat}

\subsubsection{OOD Object Detection}

For detecting OOD objects in scene-level images, unlike earlier works that constrain ID samples to a hypothetical distribution~\cite{du2022siren}, it has become a recent trend to explicitly synthesize the outlier data, and incorporate them into training to adjust models' decision boundaries.
However, due to the complexity of scene-level images, all previous works bypassed photo-realistic outlier synthesis in pixel space, and worked on generating outliers from the model's latent space~\cite{du2022vos,wu2023deep,wu2023discriminating}, adversarial attack~\cite{wilson2023safe}, or utilizing video data in the wild~\cite{du2022unknown}.
In the former, outliers can be sampled from the latent space using a simple Gaussian prior~\cite{du2022vos}, or more advanced generative models like VAE~\cite{wu2023discriminating} or diffusion model~\cite{wu2023deep}.
Yet their upperbound are commonly limited by the quality of the latent space, and synthesized samples lack interpretability.
Despite this, adversarial samples~\cite{wilson2023safe} lack semantic diversity, and auxiliary pseudo supervision from videos~\cite{du2022unknown} introduces additional requirements to the setting.
Unlike all above, our method applies LLM for OOD concept sampling, and Stable Diffusion for controllable image editing. This decouples the sampling and generation processes, elevates both parts to an unprecedented level, and achieves photo-realistic scene-level OOD image synthesis for the first time.

\vspace{-3mm}
\subsubsection{Open-world Object Detection}
The fact that real-world applications require object detectors the ability to tackle open categories is also considered in open-world object detection.
The focus of this field includes generalization to domain shifts~\cite{wang2021robust}, incremental learning of novel classes~\cite{liu2020continual,perez2020incremental,ore}, and zero-shot classification of open-vocabularies~\cite{gu2022openvocabulary,rahman2020zero}.
Meanwhile, some works~\cite{joseph2021towards,gupta2022ow} require to distinguish known objects well, be able to detect unknown objects, and finally be able to incrementally learn new objects.
These works provide different perspectives on facing the challenges of real-world applications, which complement OOD object detection as a joint effort, but are out of the scope of this paper.

\vspace{-3mm}
\subsubsection{OOD Image Classification}

Earlier paradigms for OOD image classification either post hoc adjust models' confidence score at the testing phase, or apply regularization at models' training phase.
The former line mainly focuses on the design of score functions, including confidence-based~\cite{bendale2016towards,hendrycks2017baseline,liang2018enhancing}, energy-based~\cite{liu2020energy,wang2021can,wu2023energy}, distance-based~\cite{lee2018simple,ren2021simple,sastry2020detecting,sun2022out}, gradient-based~\cite{huang2021importance}, and approximating Bayesian~\cite{miller2018dropout,miller2019evaluating,deepshikha2021monte,dhamija2020overlooked, hall2020probabilistic} methods.
The latter line of work includes regularizing models to produce lower confidence~\cite{lee2018training,hendrycks2018deep}, higher energy~\cite{liu2020energy,katz2022training}, or directly shaping latent representations~\cite{du2022siren}.
While outlier synthesis has shown to be effective by~\cite{du2022vos,tao2023non}, these are still generated in the latent space, and a parallel line study utilizing natural images~\cite{hendrycks2018deep,katz2022training,du2024does} from the wild.
Recently, photo-realistic outlier synthesis was first achieved by~\cite{du2024dream} with help from a text-conditioned diffusion model~\cite{rombach2022high}.   
However, it is not readily applicable for object detection due to the complexity of scene-level images and the requirement for object-level annotations.
In contrast, our work studies under the detection setting and requires outlier synthesis at scene level.

\vspace{-3mm}
\subsubsection{Foundation Models}

The evolution of large language models (LLMs) began with training on web-scale datasets~\cite{radford2019language,devlin2018bert}, leading to increasingly powerful foundation models~\cite{brown2020language,chowdhery2023palm} capable of harnessing vast open-world data. Notable advancements include models~\cite{achiam2023gpt} that interact with users and perform complex tasks like question answering, significantly broadening access to global knowledge. In image generation, diffusion models~\cite{ho2020denoising,rombach2022high} offer robust capabilities in synthesizing realistic content for applications such as image synthesis and inpainting. Additionally, segmentation foundation models like SAM~\cite{kirillov2023segment} represent a leap forward in precise image segmentation, benefiting from extensive data training. Foundation models provide diverse data, which provides unlimited potential for learning open-world knowledge from these models~\cite{tian2024learning}, where the effectiveness of data~\cite{tan2023data,saco2024,he2022synthetic} is also crucial to the downstreaming tasks.

\section{Method}
\label{sec:method}

\subsubsection{Preliminary} 
For OOD object detection, the training set contains only ID scene-level images $\textbf{x}^{\text{id}}$ with ID objects, annotation bounding boxes $\textbf{b}^{\text{id}}$, and semantic labels $y$, denoted as $\mathcal{D}_{\text{id}} = \left\{(\textbf{x}^{\text{id}}, \textbf{b}^{\text{id}}, \text{y}^{\text{id}})\right\}$. The labels of ID objects always belong to a close set with $K$ categories, denoted as $\text{y}^{\text{id}} \in \mathcal{Y}_{\text{id}}$ and $\mathcal{Y}_{\text{id}} = \left\{\text{y}^{\text{id}}_{1},\text{y}^{\text{id}}_{2},...,\text{y}^{\text{id}}_{K}\right\}$. During inference, for each proposed object from an input scene-level image, it is required to identify whether its category belongs to $\mathcal{Y}_{\text{id}}$ or not. 

\vspace{-3mm}
\subsubsection{Overview} 
As illustrated in \cref{fig:main}, our outlier synthesis pipeline consists of (1) synthesizing a set of effective photo-realistic scene-level OOD images $\textbf{x}^{\text{edit}}$, denoted as $\mathcal{D}_{\text{edit}} = \left\{(\textbf{x}^{\text{edit}}, \textbf{b}^{\text{edit}})\right\}$, which contains novel objects and corresponding annotation boxes $\textbf{b}^{\text{edit}}$ based on region-level editing from $\mathcal{D}_{\text{id}}$ in a fully automated, labor-free way; and (2) select and use the efficient synthetic data to provide pseudo-OOD supervisions for training OOD object detector together with the ID samples in the training set.
Further design of the pipeline requires answering the following questions: (1) how to distill the open-set knowledge embedded in foundation models to scene-level OOD data and (2) how to utilize the synthesized data to regularize the decision boundary and facilitate OOD object detection. We discuss them accordingly in \cref{subsec:method-syn} and \cref{subsec:method-use}.

\begin{figure}[t]
  \centering
  \includegraphics[width=\textwidth]{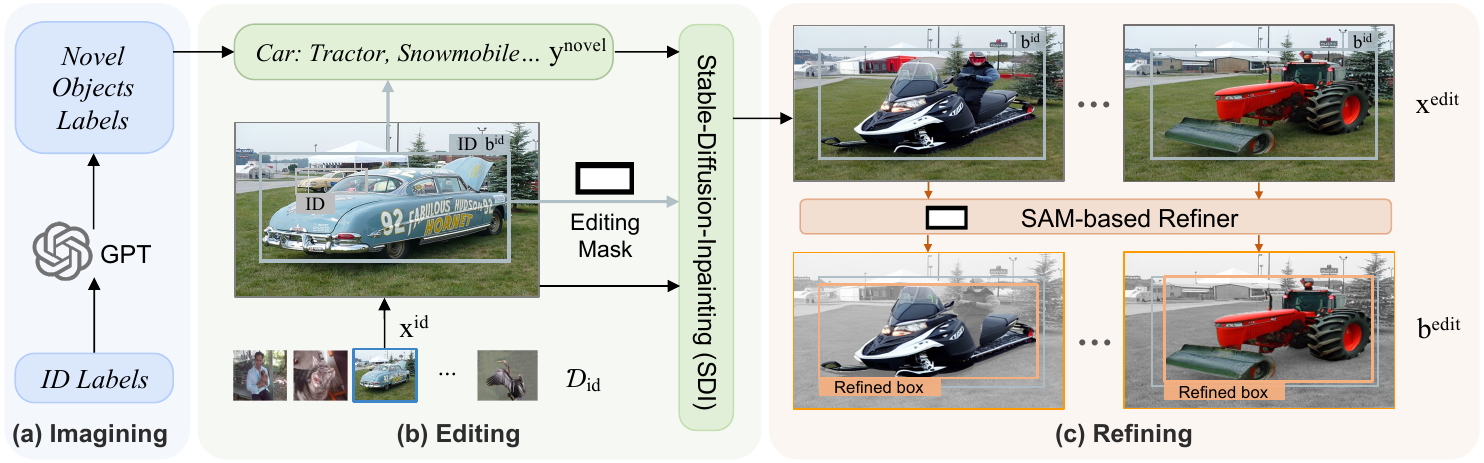}
  \caption{Detailed illustration of our outlier synthesis pipeline. It comprises (a) Instructing an LLM to imagine semantic-novel concepts given ID objects, (b) Editing the selected regions to the expected concepts via prompt-conditioned image inpainting using Stable Diffusion, and (c) Refining the bounding boxes of edited objects using SAM.}
  \label{fig:synthesize}
\end{figure}

\subsection{Synthesizing Semantic-novel Objects in Scene Images}
\label{subsec:method-syn}

\subsubsection{Imagining Novel Concepts from ID objects} 
As shown in \cref{fig:synthesize}(a), based on the ID labels $\mathcal{Y}_{\text{id}}$ in training set $\mathcal{D}_{\text{id}}$, we consider that novel concepts that are different from ID categories can be potential candidates for generating OOD objects. The next is to discover novel concepts that offer hard OOD samples sharing high visual similarity with ID samples and being contextually compatible with the scene context for object detection. Rather than relying on human labor to investigate all potential candidates, we leverage the vast knowledge and reasoning capabilities of LLM, GPT-4~\cite{achiam2023gpt} to check the visual similarity and contextual compatibility. This allows us to associate ID objects and promote the conceptualization of possible novel objects to replace existing ID objects through the use of a prompt with in-context examples~\cite{brown2020language} as:

\vspace{2mm}
\noindent
\begin{minipage}{\textwidth}
\begin{tcolorbox} 
\small
Here is a list containing several objects $\mathcal{Y}_{\text{id}}$. Now, if I provide you an object name, you should return to me objects that are similar to the usage scenario and volume of the provided object but are not in the previous object list. For example, if I give you the word: person, you should respond and only respond: `mannequin', `sculpture', `scarecrows', `doll', `puppet'.
\end{tcolorbox}
\vspace{1mm}
\end{minipage}

With its robust logical foundation and rich knowledge, the LLM envisions a collection of novel objects for each ID object label, denoted as $\mathcal{Y}_{\text{novel}}$, while maintaining the semantic separability between imagined objects and ID objects. We empirically find one in-context example that is sufficient for us to discover novel concepts. 
For each ID label $i$, we discover $M$ novel concepts using LLM $\textbf{y}^{\text{novel}}_i$ of $M$ concepts.

\vspace{-3mm}
\subsubsection{Editing Objects on Selected Regions}
With the discovered novel concepts $\mathcal{Y}_{\text{novel}} =\{\textbf{y}^{\text{novel}}_1, \textbf{y}^{\text{novel}}_2,...,\textbf{y}^{\text{novel}}_K\}$, the next step is to use them as prompts for the text-to-image generation model to generate an image. 
To generate a new image with novel concepts $y_j \in \textbf{y}^{\text{novel}}_i$, we choose to replace existing ID objects in existing images with label $y_i^{\text{id}}$ instead of finding new locations or generating one image from scratch. 
By doing so, we can ensure context compatibility and eliminate distractions from the scene context as it is preserved.  
As illustrated in \cref{fig:synthesize} (b), we use Stable-Diffusion-Inpainting~\cite{rombach2022high}, denoted as $\text{SDI}(\cdot)$, to perform region-level editing on ID images.
The ID object is denoted as $\textbf{x}^{\text{id}}$, with its corresponding annotation box $\textbf{b}^{\text{id}}$ serving as the editing mask, and the associated imagined novel concept $\textbf{y}^{\text{novel}}$ are provided as inputs to the $\text{SDI}$, which is one of the most successful models for conditional image generation and editing. Thus, an edited image $\textbf{x}^{\text{edit}}$ containing a novel object is obtained as:

\begin{equation}
\textbf{x}^{\text{edit}}=\text{SDI}(\textbf{x}^{\text{id}},\textbf{b}^{\text{id}},\textbf{y}^{\text{novel}}).
\label{eq:sdi}
\end{equation}

\vspace{-3mm}
\subsubsection{Refining Annotation Boxes of Novel Objects} Due to the randomness in diffusion models, the attributes of edited objects, such as their quality, volume, and localization, may not match the original object box. To address this issue, as depicted in \cref{fig:synthesize} (c), we design an efficient and effective refiner based on SAM \cite{kirillov2023segment} to obtain refined accurate bounding boxes on novel objects.
First, for an edited image $\textbf{x}^{\text{edit}}$ with the editing mask $\textbf{b}^{\text{id}}$, we use a padding area extended from $\textbf{b}^{\text{id}}$ as the prompt and employ SAM to output the instance mask with highest confidence $\textbf{m}^{\text{SAM}}$ for the novel object in the area:
\begin{equation}
\textbf{m}^{\text{SAM}}=\text{SAM}(\textbf{x}^{\text{edit}};\text{padding}(\textbf{b}^{\text{id}}, e)),
\label{eq:sam}
\end{equation}
where $e$ represents the range of padding. Then, we convert obtained masks $\textbf{m}^{\text{SAM}}$ to boxes $\textbf{b}^{\text{SAM}}$, and calculate IoU between $\textbf{b}^{\text{SAM}}$ and the corresponding $\textbf{b}^{\text{id}}$ to filter out novel objects that vary highly in scale:
\begin{equation}
\left\{\textbf{b}^{\text{edit}}\right\}=\left\{\left.\textbf{b}^{\text{SAM}}\middle|\right.\text{IoU}(\textbf{b}^{\text{SAM}},\textbf{b}^{\text{id}})>\gamma\right\},
\label{eq:iou}
\end{equation}
where $\gamma$ denotes a threshold value on IoU. It ensures a high enough recall rate to rule out the instability of Stable Diffusion and SAM and uncontrollable localization of the edited objects. Thus we obtain the synthetic outlier data $\mathcal{D}_{\text{edit}}$ as illustrated in \cref{fig:main}.

\subsection{Mining Hard OOD Samples and Model Training}
\label{subsec:method-use}

\subsubsection{Mining Hard OOD Objects with High Visual Similarities for Training} We consider the novel objects that are most likely to be confused with the corresponding ID objects by the object detector as the most effective ones. We thus aim to find synthetic OOD samples that are most easily confused as ID to participate in training based on pairwise similarity in the latent space of the pre-trained object detector. For each novel object with bounding box $\textbf{b}^{\text{edit}}$ in the synthetic data $\mathcal{D}_{\text{edit}}$, we construct it with the corresponding original ID object with its bounding box as a pair: $\left\{(\textbf{b}^{\text{edit}},\textbf{x}^{\text{edit}}),(\textbf{b}^{\text{id}}, \textbf{x}^{\text{id}})\right\}$. For an off-the-shelf object detector, denoted by $\mathcal{F}_\text{det}$, we extract latent features, $\textbf{z}^{\text{edit}}$ and $\textbf{z}^{\text{id}}$, for each pair: 
\begin{equation}
\textbf{z}^{\text{edit}},\textbf{z}^{\text{id}}=\mathcal{F}_\text{det}(\textbf{b}^{\text{edit}};\textbf{x}^{\text{edit}}),\mathcal{F}_\text{det}(\textbf{b}^{\text{id}};\textbf{x}^{\text{id}}).
\label{eq:extract}
\end{equation}
The most effective novel objects are those with visual patterns that can be easily mistaken for their corresponding ID objects by an object detector. Therefore, we filter these novel objects based on their similarity to provide pseudo-OOD supervision:
\begin{equation}
\left\{\textbf{z}^{\text{ood}}\right\}=\left\{\left.\textbf{z}^{\text{edit}}\middle|\right.\epsilon_{\textit{low}}<\text{sim}(\textbf{z}^{\text{edit}},\textbf{z}^{\text{id}})<\epsilon_{\textit{up}}\right\},
\label{eq:sim}
\end{equation}
where the similarities are computed between latent object features of edit-ID pairs. Here $\text{sim}(\cdot)$ denotes cosine similarity calculating and $\epsilon_{\textit{low}}, \epsilon_{\textit{up}}$ stand for the lower/upper similarity thresholds.

\vspace{-3mm}
\subsubsection{Optimizing ID/OOD Decision Boundary with Synthetic Samples} Once we have obtained the ID and synthetic OOD objects, we employ a lightweight MLP, denoted as $\mathcal{F}_\text{ood}$, as the OOD detector optimized with a bi-classify loss:
\begin{equation}
\mathcal{L}_\text{ood}=\mathbb{E} _{\textbf{z}\sim\textbf{z}^{\text{id}}}\left[-\log\frac{1}{1 + \exp^{-\mathcal{F}_\text{ood}(\textbf{z})}}\right]+\mathbb{E} _{\textbf{z}\sim\textbf{z}^{\text{ood}}}\left[-\log\frac{\exp^{-\mathcal{F}_\text{ood}(\textbf{z})}}{1+\exp^{-\mathcal{F}_\text{ood}(\textbf{z})}} \right].
\label{eq:optim}
\end{equation}
The aforementioned design ensures both \emph{semantic separability} and \emph{pattern similarity} for the chosen synthetic samples. As a result, our proposed method elegantly optimizes the decision boundary using only a limited number of samples. It is further demonstrated and validated in the following experiments.

\section{Experiments}
\label{sec:exp}

\subsubsection{Datasets}
Following OOD object detection setting~\cite{du2022vos}, we use the \textbf{PASCAL-VOC}~\cite{everingham2010pascal} and \textbf{Berkeley DeepDrive (BDD-100K)}~\cite{yu2020bdd100k} as the ID training datasets, which consist of 20 and 10 ID categories, respectively.
Meanwhile, we evaluate the performance of our approach on two OOD datasets: \textbf{MS-COCO}~\cite{lin2014microsoft} and \textbf{OpenImages}~\cite{kuznetsova2020open} respectively.
Categories from the OOD datasets that overlapped with the ID datasets are removed to guarantee the absence of ID categories. We report ID categories, OOD categories, and texts used in driving image synthesis in the supplementary material.

\vspace{-3mm}
\subsubsection{Metrics} 
We primarily focus on reporting the \textbf{FPR95} which represents the false positive rate of OOD samples when the true positive rate of ID samples is at 95\% and lower is better. FPR95 is widely used to assess the OOD object detection performance~\cite{du2022vos,du2022siren,wu2023discriminating,wu2023deep,wilson2023safe}. Additionally, we present the Area Under the Receiver Operating Characteristic Curve (\textbf{AUROC}, higher is better) that is widely utilized to evaluate binary classification problems. Since we only train a plugin MLP on top of existing object detectors for OOD detection, ID performance, \eg, mean Average Precision (mAP) is unchanged and thus omitted.

\vspace{-.3cm}
\subsubsection{Implementation Details}
For our synthetic data, in order to maintain the experimental setting of OOD object detection and avoid leaking prior knowledge of foundation models, we \textbf{remove} all imagined novel objects that have the same or similar meaning as the ground truth OOD data categories. For model training, we follow the architectures of the baseline method~\cite{du2022vos,wilson2023safe} to use a Faster R-CNN~\cite{faster-rcnn} as the base object detector with ResNet-50~\cite{he2016deep} firstly. Then we trained a simple and lightweight 3-layer MLP. We follow~\cite{wilson2023safe} to extract multi-level features as training samples. ID samples are extracted from ID images and OOD samples are extracted from selected synthesized images with OOD bounding boxes. For training on the PASCAL-VOC dataset, we employ a learning rate of 1e-4, while for the BDD-100K dataset, we use a learning rate of 5e-5. Both training processes utilize a momentum of 0.9, a dropout rate of 0.5, and a batch size of 32. All training is conducted on GeForce RTX 3090 GPUs.

\subsection{Main Results on OOD Object Detection}

\begin{table}[t]
\centering
\renewcommand{\arraystretch}{1.1}
\caption{Comparing on varied ID (PASCAL VOC~\cite{everingham2010pascal}, BDD-100K~\cite{yu2020bdd100k}) and OOD (MS-COCO~\cite{lin2014microsoft}, OpenImages~\cite{kuznetsova2020open}) datasets, our method significantly outperforms other methods on different metrics and achieves SOTA performance on OOD object detection. Our method is validated on two different existing object detectors, Faster R-CNN~\cite{faster-rcnn} and VOS~\cite{du2022vos} (denoted as \emph{Ours}$_{\text{+Faster-R-CNN}}$ and \emph{Ours}$_{\text{+VOS}}$ respectively).
(Top results are shown in \textbf{bold}.)}
\label{tab:main}
\vspace{-0.1cm}
\resizebox{\linewidth}{!}{ 
\begin{tabular}{lcccccccc}
    \toprule
    \multirow{3}{*}{\textbf{Method}} & \multicolumn{4}{c}{\textbf{ID:PASCAL-VOC}} & \multicolumn{4}{c}{\textbf{ID:BDD-100K}} \\
    & \multicolumn{2}{c}{\textbf{MS-COCO}} & \multicolumn{2}{c}{\textbf{OpenImages}} & \multicolumn{2}{c}{\textbf{MS-COCO}} & \multicolumn{2}{c}{\textbf{OpenImages}} \\
    \cmidrule(lr){2-3} \cmidrule(lr){4-5} \cmidrule(lr){6-7} \cmidrule(lr){8-9}
    & \textbf{FPR95}$\downarrow$ & \textbf{AUROC}$\uparrow$ & \textbf{FPR95}$\downarrow$ & \textbf{AUROC}$\uparrow$ & \textbf{FPR95}$\downarrow$ & \textbf{AUROC}$\uparrow$ & \textbf{FPR95}$\downarrow$ & \textbf{AUROC}$\uparrow$\\
    
    \midrule
    MSP~\cite{hendrycks2017baseline} & 70.99 & 83.45 & 73.13 & 81.91 & 80.94 & 75.87 & 79.04 & 77.38 \\
    ODIN~\cite{liang2018enhancing} & 59.82 & 82.20 & 63.14 & 82.59 & 62.85 & 74.44 & 58.92 & 76.61 \\
    Mahalanobis~\cite{lee2018simple} & 96.46 & 59.25 & 96.27 & 57.42 & 57.66 & 84.92 & 60.16 & 86.88 \\
    Energy score~\cite{liu2020energy} & 56.89 & 83.69 & 58.69 & 82.98 & 60.06 & 77.48 & 54.79 & 79.60 \\
    Gram matrices~\cite{sastry2020detecting} & 62.75 & 79.88 & 67.42 & 77.62 & 60.93 & 74.93 & 77.55 & 59.38 \\
    Generalized ODIN~\cite{hsu2020generalized} & 58.57 & 83.12 & 70.28 & 79.23 & 57.27 & 85.22 & 50.17 & 87.18 \\
    CSI~\cite{tack2020csi} & 59.91 & 81.83 & 57.41 & 82.95 & 47.10 & 84.09 & 37.06 & 87.99 \\
    GAN-synthesis~\cite{lee2018training} & 60.93 & 83.67 & 59.97 & 82.67 & 57.03 & 78.82 & 50.61 & 81.25 \\
    SIREN-KNN~\cite{du2022siren} & 47.45 & 89.67 & 50.38 & 88.80 & - & - & - & -\\
    VOS~\cite{du2022vos} & 47.53 & 88.70 & 51.33 & 85.23 & 44.27 & 86.87 & 35.54 & 88.52 \\
    SR-VAE~\cite{wu2023discriminating} & 42.17 & 90.28 & 46.26 & 87.89 & 32.23 & 90.69 & 21.81 & 93.55 \\
    DFDD~\cite{wu2023deep} & 41.34 & \textbf{90.79} & 44.52 & 88.65 & 30.71 & 90.74 & 22.67 & 92.48 \\
    SAFE~\cite{wilson2023safe} & 47.40 & 80.30 & 20.06 & 92.28 & 32.56 & 88.96 & 16.04 & 94.64 \\
    \cellcolor{mygray}\emph{Ours}$_{\text{+Faster-R-CNN}}$ & \cellcolor{mygray}\textbf{36.44} & \cellcolor{mygray}86.52 & \cellcolor{mygray}\textbf{13.34} & \cellcolor{mygray}\textbf{95.37} & \cellcolor{mygray}\textbf{22.67} & \cellcolor{mygray}\textbf{95.44} & \cellcolor{mygray}\textbf{12.96} & \cellcolor{mygray}\textbf{96.26} \\
    \cellcolor{mygray}\emph{Ours}$_{\text{+VOS}}$ & \cellcolor{mygray}\textbf{34.97} & \cellcolor{mygray}87.90 & \cellcolor{mygray}\textbf{11.25} & \cellcolor{mygray}\textbf{96.96} & \cellcolor{mygray}\textbf{23.09} & \cellcolor{mygray}\textbf{94.32} & \cellcolor{mygray}\textbf{14.12} & \cellcolor{mygray}\textbf{96.41} \\
    \bottomrule
\end{tabular}}
\end{table}

We evaluate the performance of the proposed method on different challenging benchmarks and obtain notable results (see \cref{tab:main}). As the first work to introduce synthetic scene-level natural images as OOD samples, we incorporate our data-centric method to two off-the-shelf object detectors~\cite{faster-rcnn,du2022vos}, achieving new state-of-the-art performance in OOD object detection. 

Compared with previous methods, we present comprehensive and substantial performance improvements on FPR95. The encouraging outcomes clearly show that our synthetic data offers superior OOD supervision and are well-suited for forming a precise decision boundary between ID and OOD samples as illustrated in \cref{fig:main}, which significantly reduces the interference caused by contextual information when optimizing the decision boundary.

Bridged by our synthetic data, foundation models' extensive knowledge and powerful logic about novel concepts are effectively injected into our model through novel concept imagining and region-level editing. 
Furthermore, powered by the similarity-based filter, our synthetic data proves to be highly effective. Compared with SAFE~\cite{wilson2023safe} which uses a similar framework, we only use around 25\% (on PASCAL-VOC) and 20\% (on BDD-100K) of auxiliary data to achieve a significant performance improvement. Further analysis is presented in \cref{subsec:exp-data}.

\subsection{Ablation Study}
\label{subsec:exp-data}

\begin{table}[t]
\centering
\renewcommand{\arraystretch}{1.2}
\caption{Ablation on the number of our synthetic data in training. Taking PASCAL-VOC as the ID dataset, we perform seven groups of random sampling with different numbers in the synthetic dataset to extract features as OOD samples, evaluate and report the performance on the MS-COCO/OpenImages datasets.}
\label{tab:ablation_feature_num}
\vspace{-0.1cm}
\tablestyle{3pt}{1.05}
\resizebox{\linewidth}{!}{ 
\begin{tabular}{lccccccc}
    \toprule
    \textbf{\#Sample} & \textbf{14k} & \textbf{12k} & \textbf{10k} & \textbf{8k} & \textbf{6k} & \cellcolor{mygray}\textbf{4k} & \textbf{2k}\\
    \midrule
    \textbf{FPR95}$\downarrow$ & 36.70/12.96 & 36.27/13.01 & 37.31/13.25 & 36.53/13.30 & 36.70/12.96 & \cellcolor{mygray}36.44/13.34 & 37.82/13.87\\
    \textbf{AUROC}$\uparrow$ & 86.65/95.54 & 86.68/95.55 & 86.64/95.51 & 86.69/95.52 & 86.75/95.56 & \cellcolor{mygray}86.52/95.44 & 86.03/95.18\\
    \bottomrule
\end{tabular}}
\vspace{-0.3cm}
\end{table}

\subsubsection{Number of Training Samples}
We conduct an extensive ablation study on the quantity of synthetic data utilized, as illustrated in \cref{tab:ablation_feature_num}. We employ seven sets of synthetic data with varying quantities as OOD samples, using PASCAL-VOC as the ID dataset. Features are extracted and the OOD detector is trained based on the same Faster R-CNN checkpoint for each set. It is noteworthy that as the number of samples decreases from 14k to 2k, the performance does not deteriorate but rather maintains stable and superior results. This highlights our method's data efficiency (SAFE used 16k samples). Combined with our feature similarity-based filtering strategy as in \cref{subsec:method-use}, a small number of high-quality OOD samples with \emph{visual similarity} directly promotes the optimization of precise decision boundaries to achieve stable improvements.

Meanwhile, we use the same detector checkpoint to assess the parallel baseline, SAFE~\cite{wilson2023safe}, and get the performance of \textbf{50.86/23.60} on \textbf{FPR95} and \textbf{78.15/91.42} on \textbf{AUROC}. SAFE augments about 16k images and extracts more than 100k instance-level features as OOD samples. In contrast, our approach utilizes fewer synthetic images and extracts only one instance-level feature from the edited novel object in each synthetic image as an OOD sample, resulting in significantly superior performance compared to SAFE~\cite{wilson2023safe}.

\vspace{-.3cm}
\subsubsection{Number of Concepts to Imagine}

We employ in-context learning to guide LLM in associating new objects for driving image editing.
For each ID obejct, the LLM connects a steady stream of novel objects. We further explore the impact of the number of corresponding novel objects for each ID object on data performance. We randomly sample different numbers of novel objects from LLM's responses for each ID object.
As shown in \cref{tab:ablation_concept_num}, the performance remains stable despite changes in the number of concepts, further highlighting the stability and robustness of our synthetic data.

\begin{table}[t]
\centering
\caption{We study the impact of associating varying numbers of imagined novel objects with each ID object. Taking PASCAL-VOC as the ID dataset, we report the performance on MS-COCO/OpenImages datasets.}
\label{tab:ablation_concept_num}
\vspace{-0.1cm}
\resizebox{\linewidth}{!}{ 
\tablestyle{3pt}{1.05}
\begin{tabular}{lcccccc}
    \toprule
    \textbf{\#Sample} & \textbf{3} & \textbf{4} & \cellcolor{mygray}\textbf{5} & \textbf{6} & \textbf{7} & \textbf{8}\\
    \midrule
    \textbf{FPR95}$\downarrow$ & 36.96/13.58 & 37.13/13.15 & \cellcolor{mygray}37.31/12.82 & 36.87/13.25 & 37.91/13.53 & 37.13/13.15 \\
    \textbf{AUROC}$\uparrow$ & 86.56/95.54 & 86.63/95.51 & \cellcolor{mygray}86.46/95.47 & 86.51/95.37 & 86.43/95.35 & 86.44/95.56 \\
    \bottomrule
\end{tabular}}
\end{table}

\begin{table}[t]
\centering
\renewcommand{\arraystretch}{1.1}
\caption{We randomly sample the same numbers of OOD features as the main experiment instead of using the feature filter (denoted as w/o filter), and evaluate on multiple datasets. The obtained results demonstrate the effectiveness of the proposed data filter.}
\label{tab:ablation_feature_filter}
\vspace{-0.1cm}
\resizebox{\linewidth}{!}{ 
\begin{tabular}{lcccccccc}
    \toprule
    \multirow{3}{*}{\textbf{Method}} & \multicolumn{4}{c}{\textbf{ID:PASCAL-VOC}} & \multicolumn{4}{c}{\textbf{ID:BDD-100K}} \\
    & \multicolumn{2}{c}{\textbf{MS-COCO}} & \multicolumn{2}{c}{\textbf{OpenImages}} & \multicolumn{2}{c}{\textbf{MS-COCO}} & \multicolumn{2}{c}{\textbf{OpenImages}} \\
    \cmidrule(lr){2-3} \cmidrule(lr){4-5} \cmidrule(lr){6-7} \cmidrule(lr){8-9}
    & \textbf{FPR95}$\downarrow$ & \textbf{AUROC}$\uparrow$ & \textbf{FPR95}$\downarrow$ & \textbf{AUROC}$\uparrow$ & \textbf{FPR95}$\downarrow$ & \textbf{AUROC}$\uparrow$ & \textbf{FPR95}$\downarrow$ & \textbf{AUROC}$\uparrow$\\
    
    \midrule
    w/o filter & 39.29 & 85.46 & 13.68 & 95.22 & 25.45 & 93.17 & 15.32 & 95.83\\
    \cellcolor{mygray}\emph{Ours} & \cellcolor{mygray}\textbf{36.44} & \cellcolor{mygray}\textbf{86.52} & \cellcolor{mygray}\textbf{13.34} & \cellcolor{mygray}\textbf{95.37} & \cellcolor{mygray}\textbf{22.67} & \cellcolor{mygray}\textbf{95.44} & \cellcolor{mygray}\textbf{12.96} & \cellcolor{mygray}\textbf{96.26} \\
    \bottomrule
\end{tabular}}
\vspace{-0.3cm}
\end{table}

\vspace{-0.4cm} \subsubsection{SAM-based Refiner}
\label{subsubsec:sam}
As mentioned in \cref{subsec:method-syn}, we propose to utilize SAM-based refiner to correct the bounding boxes of novel objects to obtain higher-quality instance-level OOD features. Therefore, we comparatively remove the proposed refiner and directly used the corresponding editing masks as bounding boxes to extract OOD features for training. Taking PASCAL-VOC as the ID dataset, after removing the refiner, we obtain \textbf{39.55/13.72} of \textbf{FPR95} and \textbf{85.94/95.37} of \textbf{AUROC} on MS-COCO/OpenImages datasets, which is better than previous methods but worse than the results (Faster R-CNN + \emph{Ours} in \cref{tab:main}) when using the refiner. This proves that OOD supervision signals contained in the synthetic data are already extracted and achieve good results under the localization of the fuzzy boxes, but more precise boxes mean higher quality features. More demos and analyses of the SAM-based refiner are shown in the supplementary material.

\begin{figure}[tb]
  \centering
  \includegraphics[width=\textwidth]{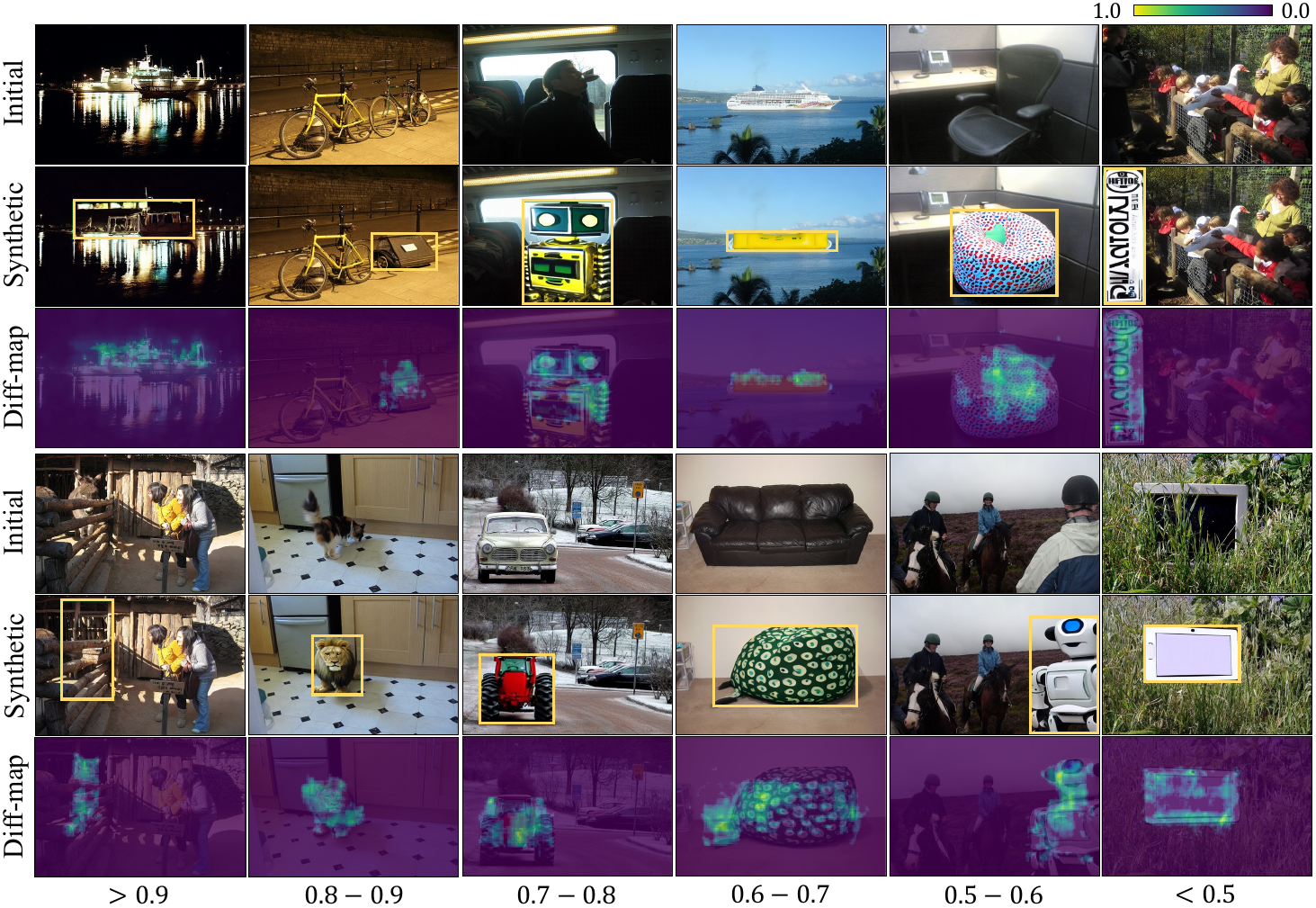}
  \vspace{-0.4cm}
  \caption{We show cases on six intervals of feature similarity (consistent with \cref{eq:sim}, indicated at the bottom of the figure). The first line contains the corresponding initial images, the second line contains the synthetic images with the corresponding boxes of the novel objects (yellow boxes), and the third line contains the difference heat maps of the latent feature maps extracted from the above image pairs (superimposed on the corresponding synthetic images, denoted as $\text{Diff-map}$).}
  \label{fig:simlarity}
  \vspace{-0.4cm}
\end{figure}

\vspace{-0.4cm} \subsubsection{Similarity-based Filter} The filter is designed to incorporate the most useful data into training, and avoid unnecessary noise. The design is reflected in two aspects: on one hand, the outlier object should process similar visual patterns to the original object, thus being confusing and can facilitate learning; on the other hand, over-high similarity may indicate failures of the editing process (\eg, when the concept is not an object).
These considerations are applied as thresholding on pairwise cosine similarity between object features, as in \cref{eq:sim}.
As shown in \cref{tab:ablation_feature_filter}, this filter brings a notable improvement across benchmarks.

To provide more insights on the choices of filtering thresholds, we display some cases in different intervals of feature similarity in \cref{fig:simlarity}. We show the synthetic images, the corresponding initial images, and the difference between feature maps (denoted as Diff-map), respectively. The Diff-maps prove that the edited area is sensitively attended to by the model.
But for images with extremely high similarity ($>0.9$), they always contain some editing failures and blurs. As illustrated on the top of the first column in \cref{fig:simlarity}, it is not intuitively evident what the \textit{ship} has been edited into (the target object is a \textit{raft}).
Besides, as the similarity upperbound decreases, we progressively obtain more realistic and reasonable images. But note that when the similarity is excessively low, as seen in the last column of \cref{fig:simlarity}, the objects are edited into the corresponding text or an unnatural object, leading to image distortion. This strongly supports the idea that the quality and usability of edited images are closely connected to visual similarity. More cases and analyses are presented in the supplementary material.

\subsection{Discussions on Outlier Synthesis}

\begin{table}[t]
\centering
\renewcommand{\arraystretch}{1.1}
\caption{Comparing varied images as OOD samples for training, we first show some synthetic object-centric images generated by Stable Diffusion (left side). Then with PASCAL-VOC as ID dataset, we report the results obtained by using synthetic object-centric images (denoted as object-centric images) and scene-level images with novel objects but without bounding boxes (denoted as scene-level w/o boxes) as OOD samples to participate in training.}
\label{tab:ablation_natural_image}
\vspace{-0.2cm}
\resizebox{\linewidth}{!}{ 
\begin{tabular}{cccccc}
    \toprule
     \textbf{Object-centric Images} &\multirow{2}{*}{\textbf{Data}} & \multicolumn{2}{c}{\textbf{MS-COCO}} & \multicolumn{2}{c}{\textbf{OpenImages}} \\ 
    \cmidrule(lr){3-4} \cmidrule(lr){5-6}
    \multirow{4}{*}{ \includegraphics[height=1.8cm]{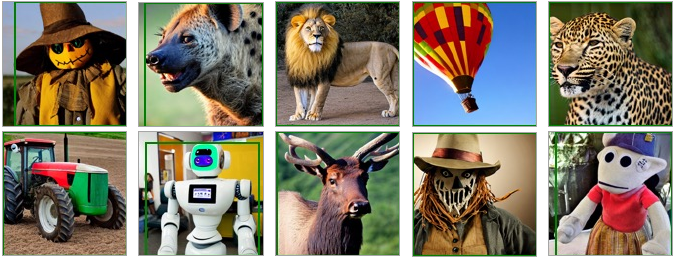}}& & \textbf{FPR95}$\downarrow$ & \textbf{AUROC}$\uparrow$ & \textbf{FPR95}$\downarrow$ & \textbf{AUROC}$\uparrow$\\
    
    \cmidrule{2-6}

    &object-centric images & 51.99 & 81.48 & 20.70 & 93.85\\
    &scene-level w/o boxes & 48.01 & 82.38 & 18.61 & 93.44\\
    &\cellcolor{mygray}\emph{Ours} & \cellcolor{mygray}\textbf{36.44} & \cellcolor{mygray}\textbf{86.52} & \cellcolor{mygray}\textbf{13.34} & \cellcolor{mygray}\textbf{95.37}\\
    \bottomrule
\end{tabular}}
\vspace{-0.4cm}
\end{table}

\subsubsection{Scene-level Editing Matters} Through regional-level editing, we replace the ID object with a novel object with a bounding box and ensure consistent context information. However, some simpler methods based on foundation models also achieve the acquisition and use of OOD data. For example, Dream-OOD~\cite{du2024dream} uses well-trained text-conditional space and diffusion model to synthesize realistic object-centric data for promoting OOD image classification. Similarly, keeping other settings unchanged, we use our novel concepts to drive Stable-Diffusion instead of Stable-Diffusion-Inpainting~\cite{rombach2022high} to synthesize novel images, which are also processed and filtered by our proposed refiner and filter (some synthesized images are shown in \cref{tab:ablation_natural_image}), thereby participating in the training as OOD supervision. However, as shown in \cref{tab:ablation_natural_image} (object-centric images), the synthetic novel object-centric images do not aid in training and result in poor performance, even though they possess high visual quality. This clearly validates our decision to edit scene-level images rather than composing new ones, and highlights the significance of maintaining contextual consistency.

Additionally, we examine the possibility of using the edited scene-level image as a whole (ignoring the boxes) as OOD samples in the training process.
The results, as depicted in \cref{tab:ablation_natural_image} (scene-level w/o boxes), are significantly inferior compared to our method's performance. This demonstrates that controllable bounding boxes are indispensable in this task.

\begin{figure}[tb]
  \centering
  \includegraphics[width=\textwidth]{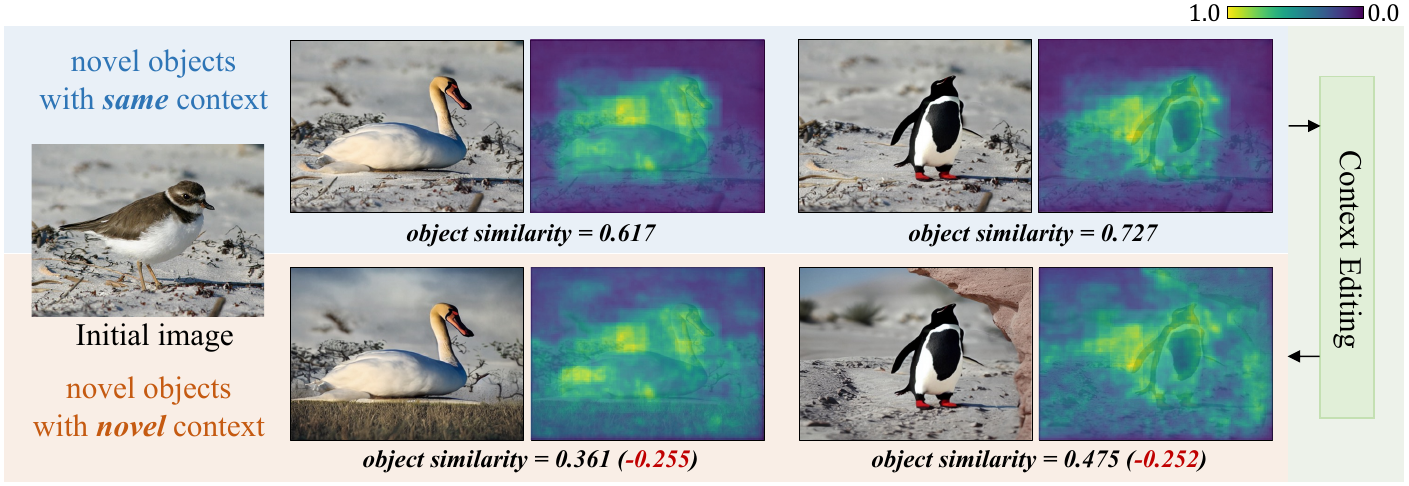}
  \vspace{-0.3cm}
  \caption{We edit the context of the synthetic data (in blue box) so that the images contain novel objects and novel context (in orange box). Then we calculate the similarity between the instance-level feature of the corresponding objects from all synthetic images and the instance-level feature in the initial image (left, the bird). The similarities and the corresponding difference maps are shown in the figure.}
  \label{fig:background}
  \vspace{-0.4cm}
\end{figure}

\vspace{-0.3cm} \subsubsection{Context Consistency Matters} Given the importance of scene-level synthesis as discussed in the previous paragraph, we study the factors that make scene-level editing indispensable, and find context consistency to be a crucial one.
Besides calculating the similarity between ID/OOD object pairs before/after box-conditioned editing, we also try further editing parts of the background of the already edited images, and also calculate its object similarity with the initial object.
As shown in \cref{fig:background}, even small editing on parts of the background (out of the object boxes) can make foreground objects `look' notably different as perceived by the detector.
This highlights the importance of keeping the context unchanged when synthesizing outlier samples, in that if the context is changed, the model easily identifies the object as OOD and cannot break the context bias.

\section{Discussion}
\label{sec:discuss}

\vspace{-0.1cm} \subsubsection{What Type of OOD Data Matters?} 
We are the first to explore how to edit scene-level images to include novel categories, which contain annotation boxes and ensure context consistency, facilitating OOD object detection. 
The achieved state-of-the-art performance (\cref{tab:main}) benefits from the optimization of decision boundaries driven by high-quality OOD features.
Our exploration demonstrates that \textit{annotation boxes} and \textit{context consistency} are particularly important for synthesizing high-quality OOD instances.
On the one hand, high-quality annotation boxes provide the possibility to extract high-quality instance-level features from scene-level images, while unrefined boxes (\cref{subsubsec:sam}) or discarded boxes (\cref{tab:ablation_natural_image}) will have a negative impact on performance.
On the other hand, context consistency ensures the most effective OOD features are not interfered by different contexts and selected for utilizing (\cref{fig:background}).

\vspace{-.3cm}
\subsubsection{How to Synthesize Suitable OOD Data?} 
We are the first to build \emph{an automatic, transparent, and low-cost pipeline} (\cref{subsec:method-syn}) for synthesizing scene-level images containing novel objects with annotation boxes and context consistency.
It benefits object detectors' robustness and reliability to unseen data and sets up clear state-of-the-art on multiple OOD object detection benchmarks.
Specifically, we organically combine and cleverly use different foundation models~\cite{achiam2023gpt,rombach2022high,kirillov2023segment} (\cref{subsec:method-syn}) to distill open-world knowledge and inpaint the existing scenes for simulating real OOD scenarios. 
In addition, our design \emph{takes into account the instability of the current foundation models} and can release better potential performance in the future development of foundation models.

\vspace{-.2cm}
\subsubsection{How to Select Suitable OOD Data?} We are the first to \textit{explicitly decouple OOD data synthesis and selection}. On the one hand, we ensure the separability of the synthetic objects in semantic concepts through open-world knowledge provided by LLMs (\cref{subsec:method-syn}). On the other hand, we ensure the similarity of ID and OOD objects in visual patterns (\cref{subsec:method-syn}), thereby optimizing the precise decision boundary. 
This line of thinking has the potential to facilitate more open-world solutions.

\vspace{-.2cm}
\subsubsection{Broader Impacts}
Beyond showcasing engineering success via effectively combining specific foundation models, our work uncovers the untapped potential of the text-to-image generative models and visual foundation models in pushing forward the OOD object detection task to effectively leverage the off-the-shelf open-world data knowledge~\cite{kong2021opengan,zheng2023out,chang2024matters}.
More importantly, our work establishes a bridge between OOD object detection and the latest advancements in deep learning, enabling it to benefit from ongoing developments, go beyond isolated academic practice, and resolve practical challenges in open-world applications. 
Meanwhile, automating novel data generation and curation will inspire more tasks in more modalities, such as in visual-language~\cite{radford2021learning,wen2024generalization} and 3D vision~\cite{liu2023mars3d,wu2024cl,deng2024can}.

\section{Conclusion}
\label{sec:conclu}

In this paper, we investigate improving OOD object detection by distilling open-world data knowledge from text-to-image generative models. We develop an automatic and cost-effective data curation pipeline, SyncOOD, that leverages foundation models as tools to obtain meaningful open-set data from generative models. Through extensive studies, we discover that object boxes and context consistency of the generated data contribute to the improvement of OOD object detection performance.
Our comprehensive experiments demonstrate that SyncOOD not only advances the state-of-the-art in OOD object detection but also emphasizes the untapped potential of utilizing large-scale generative models for enhancing the robustness of machine learning systems in open-world settings. As an initial exploration in leveraging foundation models for OOD object detection, we hope our promising results encourage further research in advancing this area in the future.

\clearpage  

%
%

\section*{Acknowledgments}

This work has been supported by Hong Kong Research Grant Council - Early Career Scheme (Grant No. 27209621), General Research Fund Scheme (Grant No. 17202422), Theme-based Research (Grant No. T45-701/22-R) and RGC Matching Fund Scheme (RMGS). Part of the described research work is conducted in the JC STEM Lab of Robotics for Soft Materials funded by The Hong Kong Jockey Club Charities Trust. We would like to thank Chirui Chang, Xiaoyang Lyu, Haoru Tan, and Xiuzhe Wu for their insightful discussions.

\bibliographystyle{splncs04}
\bibliography{egbib}
\end{document}